 \documentclass[final]{cvpr}

\usepackage{times}
\usepackage{epsfig}
\usepackage{graphicx}
\usepackage{amsmath}
\usepackage{amssymb}

\usepackage[mathscr]{eucal}
\usepackage[cal=boondox]{mathalfa}

\usepackage[pagebackref=true,breaklinks=true,colorlinks,bookmarks=false]{hyperref}

\def\cvprPaperID{7593} 
\def\confYear{CVPR 2021}

\graphicspath{{figs/}}
\usepackage{marvosym}
\usepackage{subfig}
\usepackage{booktabs}
\usepackage{multirow}
\usepackage[linesnumbered,ruled,vlined]{algorithm2e}
\usepackage{color, soul} 

\begin{document}
	\title{LAP-Net: Adaptive Features Sampling via Learning Action Progression for Online Action Detection}
	
	\author{Sanqing Qu\textsuperscript{1},\quad Guang Chen\textsuperscript{1, 3,\Letter},\quad Dan Xu\textsuperscript{2},\quad Jinhu Dong\textsuperscript{1},\quad Fan Lu\textsuperscript{1},\quad Alois Knoll\textsuperscript{3}\\
		\textsuperscript{1}Tongji University,\ \textsuperscript{2}The Hong Kong University of Science and Technology,\\
		\textsuperscript{3}Technische Universität München\\
		{\tt\small {2011444}@tongji.edu.cn,\quad guangchen@tongji.edu.cn,\quad danxu@cse.ust.hk}\\
		{\tt\small 1651873@tongji.edu.cn,\quad lufan@tongji.edu.cn,\quad
			knoll@in.tum.de}
	}
	
	\maketitle

	\begin{abstract}
		\par Online action detection is a task with the aim of identifying ongoing actions from streaming videos without any side information or access to future frames. Recent methods proposed to aggregate fixed temporal ranges of invisible but anticipated future frames representations as supplementary features and achieved promising performance. They are based on the observation that human beings often detect ongoing actions by contemplating the future vision simultaneously. However, we observed that at different action progressions, the optimal supplementary features should be obtained from distinct temporal ranges instead of simply fixed future temporal ranges. To this end, we introduce an adaptive features sampling strategy to overcome the mentioned variable-ranges of optimal supplementary features. Specifically, in this paper, we propose a novel Learning Action Progression Network termed LAP-Net, which integrates an adaptive features sampling strategy. At each time step, this sampling strategy first estimates current action progression and then decide what temporal ranges should be used to aggregate the optimal supplementary features. We evaluated our LAP-Net on three benchmark datasets, TVSeries, THUMOS-14 and HDD. The extensive experiments demonstrate that with our adaptive feature sampling strategy, the proposed LAP-Net can significantly outperform current state-of-the-art methods with a large margin.
	\end{abstract}
	
	\begin{figure}[t]
		\centering
		\subfloat[An athlete is in the early progression of the long-jump.]{
			\begin{minipage}{0.45\textwidth}
				\centering
				\includegraphics[width=3.0in]{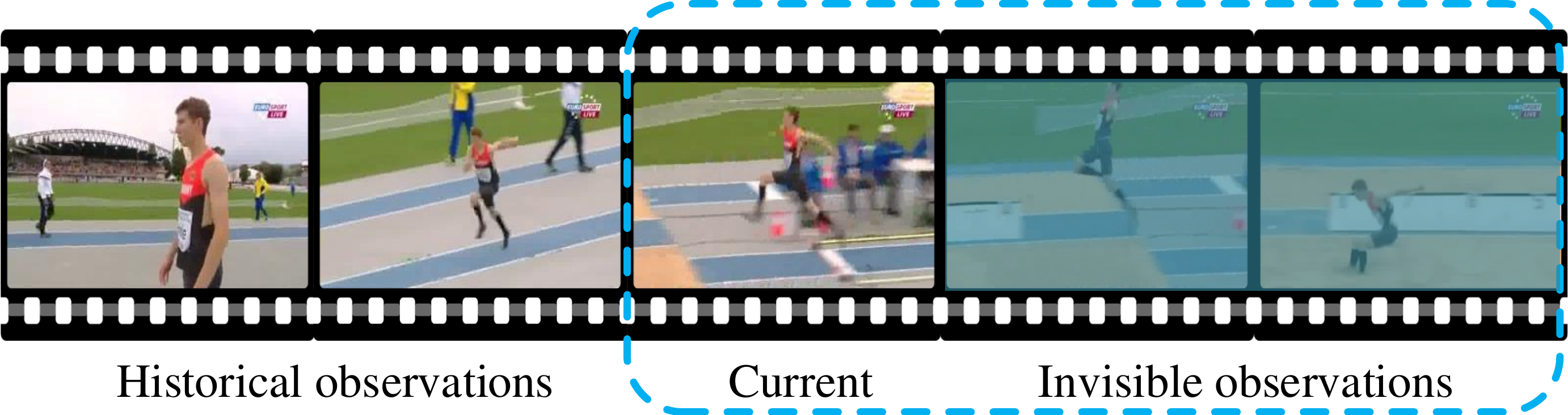}
			\end{minipage}
			\label{subfig:action_instance_at_begining_stage}}
		\\
		\subfloat[An athlete is in the middle progression of the long-jump.]{
			\begin{minipage}{0.45\textwidth}
				\centering
				\includegraphics[width=3.0in]{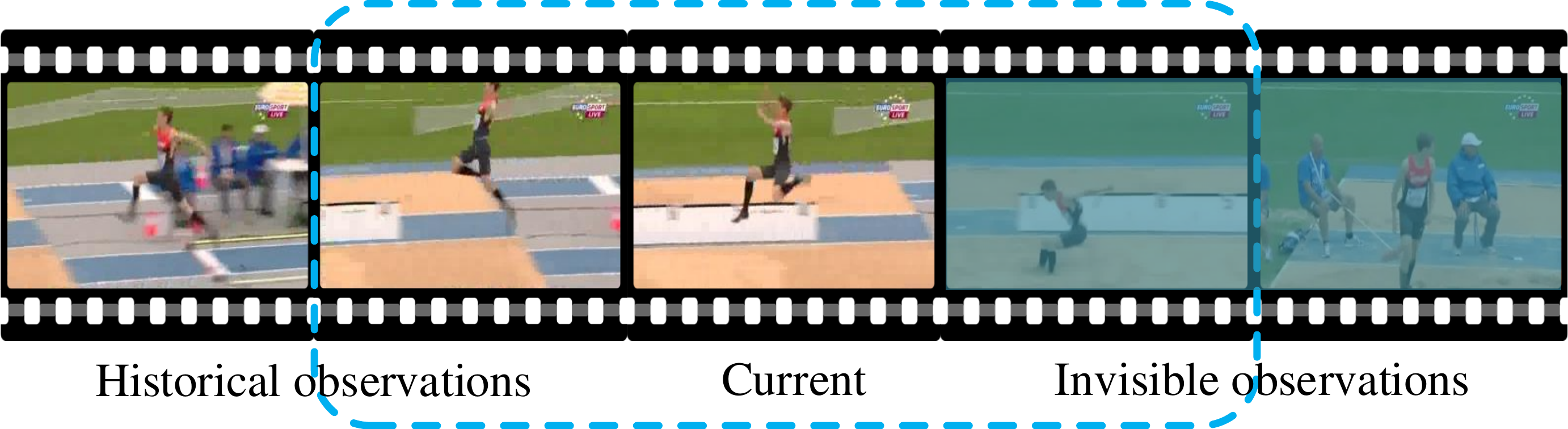}
			\end{minipage}
			\label{subfig:action_instance_at_ongoing_stage}}
		\\
		\subfloat[An athlete is in the ending progression of the long-jump.]{
			\begin{minipage}{0.45\textwidth}
				\centering
				\includegraphics[width=3.0in]{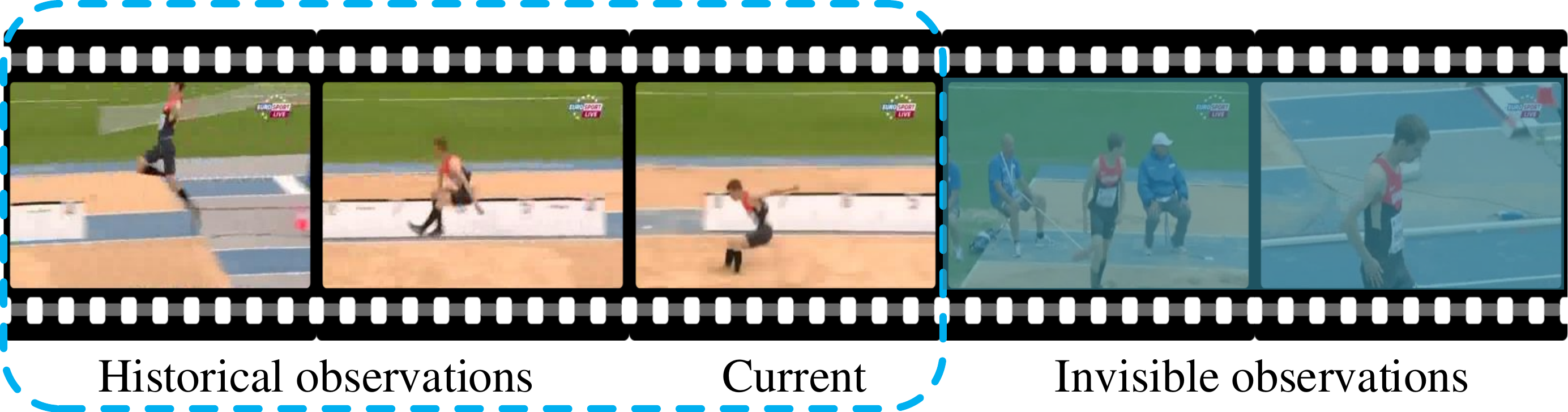}
			\end{minipage}
			\label{subfig:action_instance_at_ending_stage}}
		\caption{
			At different action progressions, the optimal supplementary representations are covered at distinct temporal ranges. For early action progression like Fig.~\ref{subfig:action_instance_at_begining_stage}, combining the predicted future representations as additional representations can help the network better understand what will happen next. As for the ending action stage like Fig.~\ref{subfig:action_instance_at_ending_stage}, 
			selecting the history observed features as supplementary representations instead of the predicted future features can help the network obtain more details and
			avoid making inaccurate decisions. (Dash line box represents the optimal temporal ranges for supplementary features.)
		}
		\label{fig:motivation}
	\end{figure}
	
	\section{Introduction}
	\par As a fundamental problem in computer vision, video understanding has attracted intense attention due to its massive potential in many fields such as video retrieval, intelligent surveillance, and behavior analysis. As one of the most challenging video understanding tasks, temporal action detection in long untrimmed videos has been well studied in an offline setting~\cite{rethinking_fastercnn, temoral_context, cascaded_boundary, BSN, BMN, 3CNet, CDC}, which allows making decisions after fully observing long videos.
	\par Different from offline temporal action detection, online action detection aims to identify ongoing actions from live video streams without any side information or access to future frames. Therefore, it is more challenging and has more real-world applications scenarios, \emph{e.g.}, autonomous driving systems, interactive virtual assistants, human robot interaction, etc. Many methods have been proposed for tackling this task~\cite{tvseries, RankLoss, online_detection_regression_1,online_detection_regression_2,lstm_action_anticipation, RED_2017, TRN_online, IDU_2020}. However, due to the lack of future information, most methods~\cite{tvseries, RankLoss,online_detection_regression_1,online_detection_regression_2,lstm_action_anticipation, IDU_2020} could not achieve the same level of detection performance as offline action detection. Inspired by the human visual cognitive system that human beings often recognize current actions by contemplating the future vision simultaneously~\cite{cognition_1, cognition_2}, several recent methods~\cite{RED_2017, TRN_online} have explored future action prediction into the online action detection task. The earlier work~\cite{RED_2017} proposed a Reinforced Encoder-Decoder (RED) network, which takes multiple observed features as input and learns to predict a sequence of future representations, and then feeds them into a classification network to recognize actions.
	A more recent work~\cite{TRN_online} extended this idea and proposed a Temporal Recurrent Network (TRN), which aggregates the estimated future representations as supplementary features and then combines them with current observations as network inputs and finally achieves better performance.
	
	\par However, we observed that the predicted future features are not always optimal for action detection and may even mislead the network to make inaccurate decisions. As shown in Fig.~\ref{subfig:action_instance_at_ending_stage}, the long-jump action is about to end but not yet over, if we aggregate the predicted future background representations as additional inputs and combine them with current observations as network inputs, the detection network is prone to be confused by the background representations. In contrast, if we retrieve history observed features as supplementary representations, the action detection network can obtain more details about current action and produces better recognition. Therefore, we argue that at different action progressions or stages, the optimal supplementary features are covered at variable temporal ranges. 
	
	\par To overcome the mentioned variable-ranges of optimal supplementary representations, we introduce an adaptive representations sampling strategy via learning action progression. As the optimal temporal range is sampled from a discrete and non-differentiable distribution parameterized by the estimation of action progression, we apply a recent Gumbel Softmax sampling approach~\cite{gumbel_softmax} to optimize this module through standard back-propagation, without introducing complex reinforcement learning as~\cite{reinforce_action_detection, reinforce_action_recognition_1, reinforce_action_recognition_2}.
	\par In this paper, we propose a novel online action detection network termed LAP-Net (Learning Action Progression Network), which integrates an adaptive supplementary features sampling strategy that first estimates current action progression and then adaptively selects the optimal temporal range for the supplementary features at each time step.  Extensive experiments on three online action detection datasets  TVSeries~\cite{tvseries}, THUMOS-14~\cite{THUMOS_14} and HDD~\cite{HDD_dataset} have demonstrated the effectiveness and general applicability of our LAP-Net. 
	\par The main contributions are summarized as follows:
	\begin{itemize}
		\item We investigate how action progression learning will affect online action detection and introduce an adaptive supplementary features sampling strategy to overcome variable-ranges of optimal supplementary representations at different action progression.
		\item We propose a novel online action detection network termed LAP-Net, which integrates a discrete adaptive supplementary feature sampling strategy. Without introducing complex reinforcement reward functions, we train the network using back-propagation through Gumbel Softmax Tricks~\cite{gumbel_softmax}.
		\item We conduct extensive experiments on three popular online action detection benchmark datasets (TVSeries~\cite{tvseries}, THUMOS-14~\cite{THUMOS_14} and HDD~\cite{HDD_dataset}) to demonstrate the superiority of our approach and obtain 0.6\%, 3.3\% and 4.3\% performance gain over current state-of-the-art methods, respectively.
	\end{itemize}
	\begin{figure*}[t]
		\centering
		\includegraphics[width=0.96\textwidth]{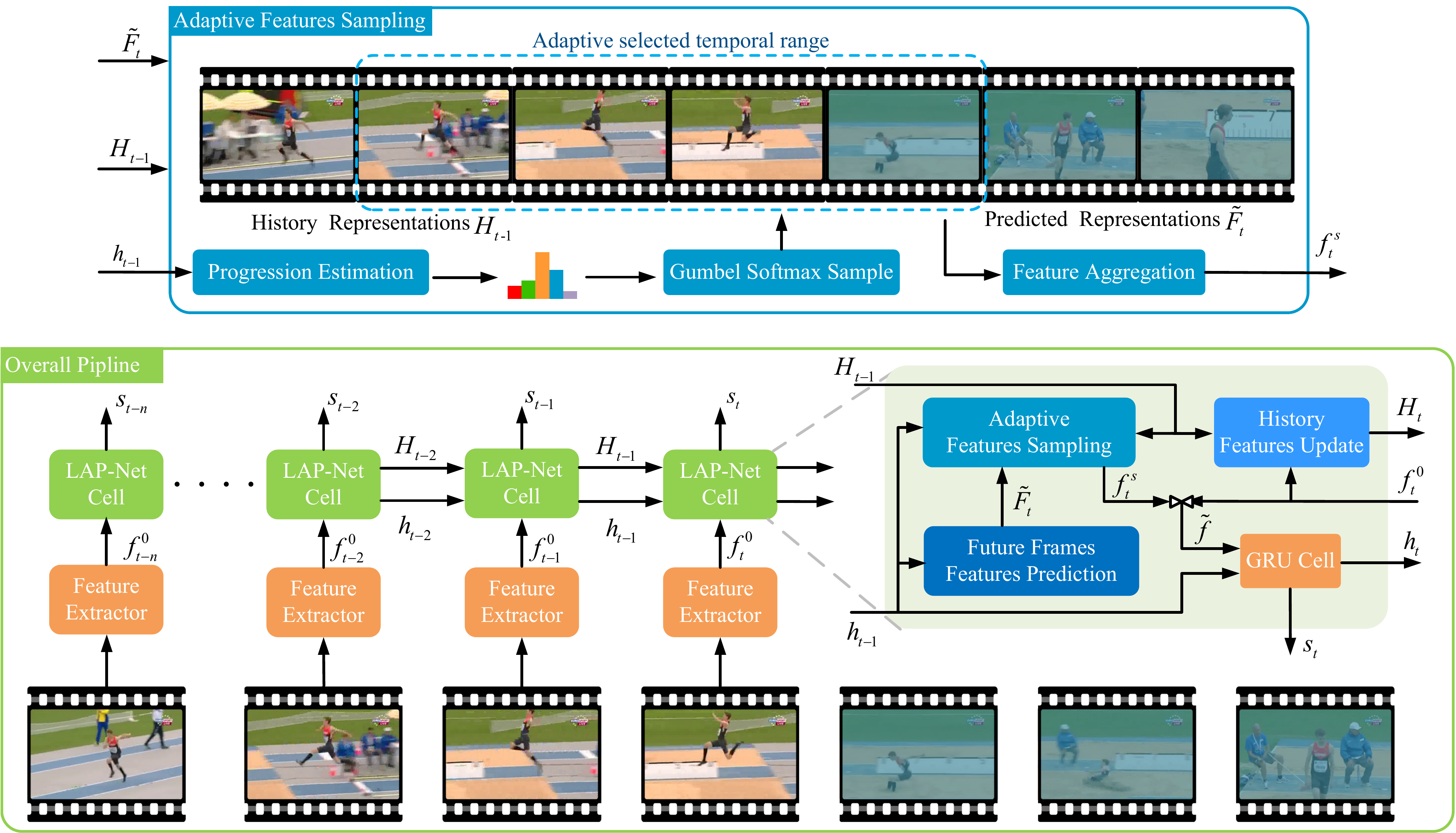}
		\caption{Framework of our proposed \emph{Learning Action Progression Network (LAP-Net)}. 
			At each time step $t$, we first utilize the future frames features prediction module to predict the invisible future representations. Then a supplementary feature sampling module is applied to aggregate the optimal supplementary features for the current action. As we argued that the optimal supplementary features are covered at variable temporal ranges at different action progression. We introduce an adaptive feature sampling strategy to overcome this limitation. Since the optimal temporal range is sampled from a discrete and non-differentiable action progression distribution, we apply the Gumbel-Softmax Tricks~\cite{gumbel_softmax} to resolve this non-differentiability, which allows us using the standard backpropagation to optimize the network during training.}
		\label{fig:LAPNet_framework}
	\end{figure*}
	
	\section{Related Work}
	\subsection{Offline Action Detection}
	\par The goal of offline action detection is to detect the start and end temporal boundaries of each action instances after fully observing long untrimmed videos. Based on the detection strategy, we briefly divided the offline action detection methods into two categories: two-stage methods and one-stage methods. The two-stage methods~\cite{CDC,RC3D, SSN,Graph_TAL,BMN} mainly follow a proposal-classification paradigm, where temporal proposals are generated first and then classified. Inspired by one-stage object detection methods, the one-stage offline action detection methods~\cite{one_stage_SSTAD, rethinking_fastercnn,single_shot_TAD, single_shot_gaussian_based} skip the proposal generation stage and directly detect action instances in long untrimmed videos. However, these methods are based on the fact that all frames are accessible, which is not possible in online action detection.
	
	\subsection{Online Action Detection}
	\par In contrast to offline action detection, online action detection aims to recognize ongoing actions from streaming video without any access to future video frames. Many approaches~\cite{tvseries, RankLoss, online_detection_regression_1,online_detection_regression_2,lstm_action_anticipation, RED_2017, TRN_online, IDU_2020} have been developed to solve this challenging task. However, due to the lack of future information, most methods~\cite{tvseries, RankLoss, online_detection_regression_1, online_detection_regression_2, lstm_action_anticipation, IDU_2020} cannot achieve comparable performance to the offline action detection. Inspired by the human visual cognitive system that human beings often recognize current actions by contemplating the future vision simultaneously~\cite{cognition_1, cognition_2}, several recent methods~\cite{RED_2017, TRN_online} have explored future action prediction into the online action detection task. An earlier work~\cite{RED_2017} proposed a Reinforced Encoder-Decoder (RED) network that takes multiple history features as input and learns to recognize and predict a sequence of future frames. A more recent work~\cite{TRN_online} extended this idea and proposed the Temporal Recurrent Network (TRN), which considers predicted future frame representations as additional input and achieves better online action detection performance. However, these methods ignore the fact that the predicted future frame representations are not always helpful for online action detection and sometimes may even mislead the network decisions. In this paper, we investigate how the optimal supplementary features could be obtained from varying ranges of frames at different action progressions. 
	
	\subsection{Adaptive Computation}
	\par With the aim to improve computational efficiency and performance, many adaptive computation approaches have been proposed~\cite{adaptive_neural_network_2017,skipnet_2018,adaptive_convolution_2018, ARNet, dynamic_zoomin_2018, RL_action_detection_2016, adaframe_2019, scsampler_2019}. These approaches can be mainly divided into adaptive computation for the most efficient neural network structure ~\cite{adaptive_neural_network_2017,skipnet_2018,adaptive_convolution_2018, ARNet} and adaptive computation for the most salient features~\cite{dynamic_zoomin_2018, RL_action_detection_2016, adaframe_2019, scsampler_2019}. 
	Though our approach is inspired by those salient feature adaptive computation methods applied in object detection and action recognition, we focus on adaptive computation in online action detection. Our goal is to adaptively aggregate the optimal supplementary features at each time step according to current action progression.
	
	\section{Method}
	\subsection{Problem Definition}
	\par Different from offline action detection that makes detection decisions after fully observing the whole video, in online action detection, there is no access to future frames or any side information.  Given a streaming video, the goal of online action detection is to identify ongoing action class $s$ in each frame, where $s \in \{1, \cdots, C\}$ and $C$ is the total number of action categories (we use index $0$ to represent the background category).
	
	\subsection{LAP-Net Framework Overview}
	\label{sec:lapnet_framework}
	\par As presented by most existing methods~\cite{tvseries, RankLoss, online_detection_regression_1,online_detection_regression_2}, it is challenging for online action detection only based on history and current action observations to recognize actions.  Inspired by human visual cognitive systems that human beings often identify current surroundings by envisioning the future vision simultaneously~\cite{cognition_1, cognition_2}, we involve future action prediction into our online action detection task to force the network to learn and contemplate the action evolving progress and better recognize current action. 
	\par Since current frame observations only contain limited information, we introduce a supplementary feature sampling module to help the network learn more discriminate representations about current frame action. Different from the previous works~\cite{RED_2017, TRN_online}, we argue that the optimal supplementary representations are covered at variable temporal ranges conditioned on different action progressions. To overcome this limitation, we integrate an adaptive sampling strategy in our feature sampling module to obtain the optimal supplementary features at distinct action progressions.
	
	\par The overall framework is presented in Fig.~\ref{fig:LAPNet_framework}. As the illustration showed, we first apply a feature extractor for each input frame to extract the individual features and then send them to the LAP-Net cell to get the recognition result. The LAP-Net cell is the core component of our network and involves adaptive features sampling, history features update, future frames features prediction, and standard GRU~\cite{GRU} based action recognition cell. 
	In the following, we will present more details about our framework. 
	
	\subsection{Adaptive Features Sampling.}
	As we present in Fig.~\ref{fig:motivation}, at different action progression, the optimal supplementary features should be obtained from distinct temporal ranges instead of fixed future temporal ranges. To this end, we introduce an action progression conditioned adaptive features sampling strategy to overcome these limitations. 
	\\
	\\
	\textbf{Action Progression Learning.}  We define the action progression space as $\Omega = \{0, 1,\cdots, P-1\}$, where $P$ is the number of predefined probable action progression state.
	At each time step $t$, we first utilize the LAP-Net cell hidden state ${h_{t-1}}$ to estimate current action progression distribution $\hat{p}_t \in \mathbb{R}^P$ over the predefined action progression space $\Omega$ through an FC layer and softmax operation. We can then directly apply a max sampling over the estimated progression distribution scores $\hat{p}_t \in \mathbb{R}^P$ to obtain the current action progression $p_s$.
	\par However, this directly sampling operation is non-differentiable, making the network unable to be optimized via the standard backpropagation. One common practice is to apply reinforcement learning and define a reward function to avoid backpropagating through these discrete samples. However, due to the undesirable fact that the reward function scales linearly with the discrete variable dimension, and it is slow to converge in many applications~\cite{reinforce_action_detection, reinforce_action_recognition_1, reinforce_action_recognition_2}.
	As an alternative, in this paper, we apply the Gumbel-Softmax Sampling~\cite{gumbel_softmax} to resolve this non-differentiability, which allows us to utilize the standard backpropagation to optimize the discrete progression estimation module in an efficient way.
	\par Specifically, at each time step $t$, after we obtained the progression distribution $\hat{p}_t \in \mathbb{R}^P$, we apply the Gumbel-Max trick~\cite{gumbel_softmax} to obtain the discrete progression $p_s$ as:
	\begin{equation}
		p_s = \mathop{\arg\max}_i (\log\hat{p}^i_t + G_i)
		\label{equ:gumbel_max_sample}
	\end{equation}
	where $G_i = -\mathop{\log(-\log U_i)}$ is a standard Gumbel distribution with $U_i$ sampled from a uniform i.i.d distribution $\mathtt{Uniform}(0, 1)$. Due to the non-differentiable property of \texttt{argmax} operation in Eq.~\ref{equ:gumbel_max_sample}, we then utilize the Gumbel-Softmax trick~\cite{gumbel_softmax} as a continuous, differentiable approximation to \texttt{argmax}. Accordingly, sampling from a Gumbel Softmax distribution allows us to backpropagate from the discrete samples to the whole network. Let $p_t$ be a one hot vector $[p_t^0, \cdots, p_t^{P-1}]$, where
	\begin{equation}
		p^i_t = \left\{\begin{array}{ll}
			1, &\text{if}\ i = p_s\\
			0, &\text{otherwise}
		\end{array}
		\right.
	\end{equation}
	The one-hot progression vector $p_t$ is relaxed to a real-valued vector $\tilde{p}_t$ using softmax:
	\begin{equation}
		\tilde{p}^i_t = \frac{\exp((\log \hat{p}^i_t + G_i)/\tau)}{\sum_{j=0}^{P-1}\exp((\log \hat{p}^j_t + G_j)/\tau)}
		\label{equ:backward_pass}
	\end{equation}
	where $\tau$ is a temperature parameter, which controls the smoothness of the progression distribution $\tilde{p}_t$. As $\tau \rightarrow 0$, $\tilde{p}_t$ becomes a one-hot vector, and as $\tau \rightarrow \infty$, $\tilde{p}_t$ converges to an uniform distribution. We set $\tau=5$ as the initial value and gradually anneal it down to 0 during training as in ~\cite{gumbel_softmax}.
	By using this Gumbel-Softmax trick, during the forward pass, we can estimate current action progression using Eq.~\ref{equ:gumbel_max_sample}. As for the backward pass, we can approximate the gradient of the discrete samples by computing the continuous softmax relaxation gradient in Eq.~\ref{equ:backward_pass}.
	\\
	\\
	\textbf{Adaptive Features Sampling.} After we obtain current action progression $p_s$, the next step is to sample and aggregate the optimal supplementary features. We denote the predicted future feature stack as $\tilde{F}_t = \{\tilde{f}_{t+i}\}^{\mathcal{l}_d - 1}_{i=0}$, where $\tilde{f}_{t+i}$ is the future $i^{\mathrm{th}}$ frame predicted representation, $\mathcal{l}_d$ is the stack size. As for the history feature stack, we denote it as $H_{t-1} = \{f^0_{t-j}\}^{\mathcal{l}_d}_{j=1}$, where $f^0_{t-j}$ is the past $j^\mathrm{th}$ frames observed representations. Note that we set the history feature stack size to $\mathcal{l}_d$, the same as the predicted future feature stack size, to keep the network with the same observation range for the history and the future. In addition, to obtain the supplementary features, the feature sampling range size is also needed, we denote it as $K$.
	\par We first concatenate the history feature stack $H_{t-1}$ and future feature stack $\tilde{F}_t$ as a feature pool $F = [H_{t-1}, \tilde{F}_t]$. Thereafter, we apply an equidistant sampling strategy to obtain the desired supplementary features $f_t^s$ from $F$ with a sampling range size of $K$. The sampling stride $s$ is set by $s =\lfloor\frac{2\mathcal{l}_d - K}{P - 1}\rfloor$. To simplify the network, we just apply a temporal average pooling strategy to aggregate these sampled representations to form the supplementary features $f_t^s$, other aggregation strategies such as max-pooling or non-local attention mechanism~\cite{non_local} are also applicable.
	
	\subsection{History and Future Features Maintaining}
	\noindent\textbf{History Features Update.}
	As we mentioned before, we maintain a history feature stack $H_t = \{f_{t-j}\}^{\mathcal{l}_d -1}_{j=0}$ to store history observed features, where $f^0_{t-j}$ is the frame-level feature at the past $j^{\mathrm{th}}$ time step. We apply a Queue mechanism to update this history feature stack. At each time step $t$, we first pop the furthest $f^0_{t-\mathcal{l}_d}$ from the feature stack $H_{t-1}$, and then push the latest feature $f^0_{t}$ to the feature stack to form the new history feature stack $H_{t}$.
	\\
	\\
	\textbf{Future Frames Features Prediction.}
	To obtain the invisible future observations, we introduce a future feature prediction module. Following to the previous work~\cite{RED_2017,TRN_online}, we apply a recurrent neural network to sequentially predict future frames representations.  We refer the interested readers to~\cite{RED_2017,TRN_online} to get more details. In order to make a good future features prediction, a future action level cross-entropy loss is introduced to this module. 
	
	\begin{equation}
		\mathscr{L}_{pre} = -\frac{1}{\mathcal{l}_d}\sum_{i=0}^{\mathcal{l}_d -1} y_{t+i} \log(\tilde{c}_{t+i})
	\end{equation}
	where $\mathcal{l}_d$ is the total steps for future feature prediction, $i$ represents the specific prediction step, $\tilde{c}_{t+i}$ is the $\mathrm{i}^\mathrm{th}$ step predicted future action classification probability, and $y_{t+i}$ is the corresponding ground truth action label. We obtain $\tilde{c}_{t+i} \in \mathbb{R}^C$ through a classification layer and a softmax operation based on the prediction unit hidden state $\tilde{h}_{t+i}$. 
	\subsection{Online Action Detection}
	To sequentially identify the streaming input video frames, we implement our LAP-Net using the gated recurrent units (GRUs)~\cite{GRU} as basic building blocks. The input to the GRU cell is the last time step hidden state $h_{t-1}$ and the concatenated feature $\tilde{f}_t = [f_t^0, f_t^s]$, where $f_t^0$ is the current frame-level features extracted by the feature extraction module, and $f_t^s$ is the supplementary features sampled through our adaptive sampling strategy. The output of this GRU cell is the updated hidden state $h_t$. 
	To identify each frame, we feed the updated hidden state $h_t$ into a classification layer to obtain the recognition score $s_t \in \mathbb{R}^C$, which is then fed into a softmax layer to get the final classification probability $c_t \in \mathbb{R}^C$. Note that, we share this classification layer with the Future Frames Features Prediction module. Same as the future action level loss calculation, we also apply the cross-entropy loss to the recognition score.
	\begin{equation}
		\mathscr{L}_{cls} = -y_{t}\log(c_t) 
	\end{equation}
	where $y_t$ is the current frame ground truth action class label.
	
	\subsection{Optimization Objective}
	\par Using the Gumbel-Softmax trick~\cite{gumbel_softmax}, we can optimize our network with standard backpropagation during training, even though the supplementary feature sample module introduces discrete progression sample operations.
	In order to ensure reliable action detection and future feature prediction, we combine the action detection loss and the future feature prediction loss as follows:
	\begin{equation}
		\mathscr{L} =\mathscr{L}_{cls} + \lambda \mathscr{L}_{pre}
	\end{equation}
	where $\lambda$ is a loss balancing factor. Note that, during training, we optimize our network as the offline action detection setting since the current and future frame labels are used. But during testing, these side information is not needed.

	\section{Experiments}
	\subsection{Experimental Setup}
	\noindent \textbf{Datasets.} We evaluate our LAP-Net on three large-scale online action detection datasets: TVSeries~\cite{tvseries}, THUMOS-14~\cite{THUMOS_14} and HDD~\cite{HDD_dataset}. TVSeries is temporally annotated with 30 realistic action categories and consists of 27 untrimmed very long videos over six popular TV series (20 videos for training and seven videos for testing). THUMOS-14 contains 1010 validation and 1574 test videos from 101 action categories. Out of these, 200 validation and 213 test videos are temporally annotated with 20 sport action categories. Following the prior works~\cite{TRN_online, IDU_2020}, we use the 200 validation videos for training and 213 test videos for evaluation.
	HDD was recorded from an autonomous driving platform and contains 137 driving session videos (nearly 104 hours). In addition to the visual data, HDD also provided non-visual vehicle CAN sensors data. Same as the prior works~\cite{TRN_online, GCN_action_recognition}, we utilize 100 sessions for training and 37 sessions for testing.
	\\
	\\
	\noindent \textbf{Implementation Details.}  Following the previous works ~\cite{RED_2017, TRN_online, IDU_2020}, we utilize a pre-trained two-stream network (TSN)~\cite{TSN_wang} to extract the frame-level features. For the TVSeries and THUMOS-14 datasets feature extraction, we first extract video frames at 24 fps and then sample 4 non-overlapping video chunks with a size of 6, i.e. a chunk consists of 0.25 seconds video frames. For each video chunk, we then apply the off-the-shelf two-stream-network pre-trained on ActivityNet-1.3~\cite{activitynet} to extract the RGB images based appearance feature and optical-flow based motion features. As for the HDD dataset, following the state-of-the-arts~\cite{HDD_dataset, TRN_online}, we first extract video frames and non-visual CAN sensor data at 3fps. Then, we apply the InceptionResNet-V2~\cite{incpetion_resnet} pre-trained on ImageNet~\cite{ImageNet} to extract visual features. For the non-visual CAN sensors data, we apply an FC layer for feature embedding.
	\par We implement our Learning Action Progression Network (LAP-Net) in Pytorch~\cite{pytorch}, and perform all the experiments with NVIDIA 2080Ti graphic cards. For the parameters tuning, we utilize the Adam optimizer~\cite{adam} with learning rate 0.0005 and weight decay 0.001. During training, we set the training sample length $\mathcal{l}_e$ to 64, the future feature prediction steps $\mathcal{l}_d$ to 8. We set the action progression state number $P$ to 4, and set the feature aggregation temporal window size $K = 7$ for THUMOS-14, $K = 9$ for TVSeries and $K = 3$ for HDD. For the GRU cell backbone, we set both the action detection cell and future feature prediction cell hidden state dimension to 4096. The training batch size is set to 16.  In addition, for data augmentation, we randomly sample chopped off $\Delta \in [1, \mathcal{l}_e]$ frames from the beginning for each epoch, and reconstruct the input video with length $l$ into $\lfloor (l-\Delta)/\mathcal{l}_e \rfloor$ discrete non-overlapping training samples. (Each training sample is consist of $\mathcal{l}_e$ consecutive input frames.) 
	\\
	\\
	\noindent \textbf{Baselines.} We compare our LAP-Net with the following existing state-of-the-art approaches.
	\begin{itemize}
		\item RED~\cite{RED_2017}, which first involves future frame prediction for online action detection and build on LSTM.
		\item TRN~\cite{TRN_online}, which proposes a temporal recurrent network (TRN) cell, where a future frames representations prediction is introduced. At each time step, the TRN cell combines predicted future representations and current observed features as network input.
		\item IDN~\cite{IDU_2020}, which proposes an Informative Discrimination Unit(IDU). At each time step, IDU iteratively aggregates the most relevant history features from a fixed history feature stack.
	\end{itemize}
	\noindent \textbf{Evaluation Metrics.} Following the most existing works, we compute  the per-frame mAP (mean average precision) to reflect the performance of online action detection. Besides, we also compute the per-frame mcAP (mean calibrated average precision) proposed in~\cite{tvseries} to evaluate the online action detection performance on TVSeries.  
	
	\subsection{Comparison with the State-of-the-Arts}
	\begin{table}[t]
		\centering
		\begin{tabular}{p{0.20\textwidth}p{0.10\textwidth}c}
			\toprule
			Method& Setting & mAP(\%)\\
			\midrule
			Two-stream CNN~\cite{two_stream_16}& \multirow{5}*{Offline} &36.2\\
			C3D + LinearInterp~\cite{CDC} & &37.0\\
			MultiLSTM~\cite{multi-lstm} & &41.3\\
			Conv\& De-conv~\cite{pred_corrective} & &41.7\\
			CDC~\cite{CDC} & &44.4\\
			\midrule
			RED~\cite{RED_2017} &\multirow{4}*{Online} &45.3\\
			TRN~\cite{TRN_online} & & 47.2\\
			IDN~\cite{IDU_2020} & &50.0\\
			\textbf{LAP-Net} (ours) & &\textbf{53.3}\\
			\bottomrule
		\end{tabular}
		\caption{Online action detection performance comparison on the THUMOS-14~\cite{THUMOS_14} dataset.}
		\label{table:thumos_14_comparison}
	\end{table}
	\begin{table}[t]
		\centering
		\begin{tabular}{p{0.16\textwidth}p{0.12\textwidth}c}
			\toprule
			Method& Inputs & mcAP(\%)\\
			\midrule
			CNN~\cite{tvseries} &\multirow{4}*{RGB}& 60.8\\ 
			RED~\cite{RED_2017} &  &71.2\\
			TRN~\cite{TRN_online} & & 75.4\\
			IDN~\cite{IDU_2020} & &76.6\\	
			\midrule
			RED~\cite{RED_2017} &\multirow{4}*{RGB + Flow}  &79.2\\
			TRN~\cite{TRN_online} & &83.7\\
			IDN~\cite{IDU_2020} & &84.7\\
			\textbf{LAP-Net} (ours) & &\textbf{85.3}\\
			\bottomrule
		\end{tabular}
		\caption{Online action detection performance comparison on the TVSeries~\cite{tvseries} dataset.}
		\label{table:tvseries_comparison}
	\end{table}
	\begin{table}[t]
		\centering
		\begin{tabular}{p{0.16\textwidth}p{0.12\textwidth}c}
			\toprule
			Method& Inputs & mAP(\%)\\
			\midrule
			CNN~\cite{tvseries} &\multirow{4}*{RGB}& 20.7\\ 
			RED~\cite{RED_2017} &  &27.2\\
			TRN~\cite{TRN_online} & &29.7\\
			\textbf{LAP-Net} (ours) & &\textbf{33.5}\\
			\midrule
			CNN~\cite{tvseries} &\multirow{4}*{RGB + CAN}& 31.3\\ 
			RED~\cite{RED_2017} &  &37.8\\
			TRN~\cite{TRN_online} & &40.8\\
			\textbf{LAP-Net} (ours) & &\textbf{45.1}\\
			\bottomrule
		\end{tabular}
		\caption{Online action detection performance comparison on the HDD~\cite{HDD_dataset} dataset.}
		\label{table:hdd_comparison}
	\end{table}
	\begin{table*}[t]
		\centering
		\scalebox{1.0}{
			\begin{tabular}{llccccccccc}
				\toprule
				\multirow{2}{*}{Dataset}&\multirow{2}{*}{Method}&\multicolumn{9}{c}{Future action prediction performance (mAP or mcAP (\%))}\\
				\cmidrule{3-11} & &0.25s &0.50s &0.75s &1.00s &1.25s &1.50s &1.75s &2.00s &Avg \\
				\midrule
				\multirow{4}{*}{THUMOS-14~\cite{THUMOS_14}}
				&ED~\cite{RED_2017}    &43.8 &40.9 &38.7 &36.8 &34.6 &33.9 &32.5 &31.6 &36.6 \\
				&RED~\cite{RED_2017}   &45.3 &42.1 &39.6 &37.5 &35.8 &34.4 &33.2 &32.1 &37.5 \\
				&TRN~\cite{TRN_online} &45.1 &42.4 &40.7 &39.1 &37.7 &36.4 &35.3 &34.3 &38.9 \\
				&LAP-Net(ours) &\bf{49.0} &\bf{47.4} &\bf{45.3} &\bf{43.2} &\bf{41.3} &\bf{39.7} &\bf{38.3} &\bf{37.0} &\bf{42.6}\\
				\midrule
				\multirow{4}{*}{TVSeries~\cite{tvseries}}
				&ED~\cite{RED_2017}    &78.5 &78.0 &76.3 &74.6 &73.7 &72.7 &71.7 &71.0 &74.5 \\
				&RED~\cite{RED_2017}   &79.2 &78.7 &77.1 &75.5 &74.2 &73.0 &72.0 &71.2 &75.1 \\
				&TRN~\cite{TRN_online} &79.9 &78.4 &77.1 &75.9 &74.9 &73.9 &73.0 &72.3 &75.7 \\
				&LAP-Net(ours) &\bf{82.6} &\bf{81.3} &\bf{80.0} &\bf{78.9} &\bf{77.9} &\bf{77.1} &\bf{76.3} &\bf{75.5} &\bf{78.7}\\
				\bottomrule
		\end{tabular}}
		\caption{Future action prediction performance comparison on the THUMOS-14~\cite{THUMOS_14} and TVSeries~\cite{tvseries} datasets.}
		\label{table:action_prediction_comparison}
	\end{table*}
	\begin{table}[t]
		\centering
		\begin{tabular}{llc}
			\toprule
			Dataset& Method & mAP/mcAP(\%)\\
			\midrule
			\multirow{2}*{THUMOS-14} & LAP-Net w/o AFS &51.3\\
			& LAP-Net w/ AFS & \textbf{53.3}\\
			\midrule
			\multirow{2}*{TVSeries} &LAP-Net w/o AFS &83.7\\
			& LAP-Net w/ AFS & \textbf{85.3}\\
			\midrule 
			\multirow{2}*{HDD} &LAP-Net w/o AFS &40.2\\
			& LAP-Net w/ AFS & \textbf{45.1}\\
			\bottomrule
		\end{tabular}
		\caption{Ablation study of effectiveness of our LAP-Net on the THUMOS-14, TVSeries and HDD datasets.}
		\label{table:ablation_study_RNN}
	\end{table}
	\noindent \textbf{Results on THUMOS-14.} We compare our LAP-Net with the state-of-the-arts on THUMOS-14 in Table~\ref{table:thumos_14_comparison}.  The results show that our LAP-Net can significantly outperform current approaches with a large margin, regardless of whether these methods are online detection based or offline detection based. Specifically, our LAP-Net achieve 3.3\% mAP performance gain over IDN~\cite{IDU_2020}, 6.1\% mAP over TRN~\cite{TRN_online} and 8.0\% mAP over RED~\cite{RED_2017}.
	\\
	\\
	\noindent \textbf{Results on TVSeries.} TVSeries is a very challenging dataset for various actions, multiple actors, unconstrained viewpoints, and a large proportion of non-action frames. We report the online action detection results in Table~\ref{table:tvseries_comparison}. Compared with the state-of-the-arts, the results show that our LAP-Net can still achieve better performance on this realistic challenging dataset. Specifically, our LAP-Net achieves 0.6\% mAP performance gain over IDN~\cite{IDU_2020}, 1.6\% performance gain over TRN~\cite{TRN_online}, and 6.1\% performance gain over RED~\cite{RED_2017}. 
	\\
	\\
	\noindent \textbf{Results on HDD.}\footnote{The IDN~\cite{IDU_2020}
		neither reported their online action detection results on the HDD dataset, nor did it open source its training source codes. So we ignore this method on the HDD dataset comparison.}
	In addition to comparing the performance of online human action detection, we also report the autonomous driving action detection performance on the HDD dataset~\cite{HDD_dataset} in Table~\ref{table:hdd_comparison}. The results show that our LAP-Net can still significantly outperform current methods. Specifically, our LAP-Net achieves 4.3\% mAP performance gain over TRN~\cite{TRN_online} and 7.3\% performance gain over RED~\cite{RED_2017}.
	\\
	\\
	\noindent \textbf{Future Action Prediction.} Besides the online action detection comparison, we also compare our LAP-Net on the future action prediction performance in Table~\ref{table:action_prediction_comparison}. Even though future action prediction is not our main task, the results show that our LAP-Net can still performs much better than TRN and the RED baselines (average mAP of 42.6\% vs. 38.9\% vs. 37.5\% on THUMOS-14 and average mcAP of 78.7\% vs. 75.7\% vs. 75.1\% on TVSeries).
	
	\subsection{Ablation Study}
	\begin{table*}[t]
		\centering
		\scalebox{0.99}{
			\begin{tabular}{llp{0.03\textwidth}p{0.04\textwidth}p{0.03\textwidth}p{0.03\textwidth}p{0.05\textwidth}p{0.03\textwidth}p{0.03\textwidth}p{0.03\textwidth}p{0.03\textwidth}p{0.03\textwidth}}		
				\toprule
				\multirow{2}*{Dataset} &\multirow{2}*{Task} &\multicolumn{5}{l}{Temporal Range Size $K$ ($P=4$)} &\multicolumn{5}{l}{Progression State Space $P$}\\
				\cmidrule(r){3-7} \cmidrule(r){8-12} & &3 &5 &7 &9 &11 &2 &3 &4 &5 &6 \\
				\midrule
				\multirow{2}*{THUMOS-14} &Online Action Detection &52.7 &53.1 &\bf{53.3} &52.6 &52.6
				&51.4 &52.7 &\bf{53.3} &53.0 &52.7 \\
				
				&Future Action Prediction &44.3 &44.1 &44.0 &44.0 &43.6
				&42.7 &44.0 &{44.0} &44.6 &44.0  \\
				\midrule
				\multirow{2}*{TVSeries} &Online Action Detection &84.7 &85.0 &85.1 &\bf{85.3} &84.9
				&85.0 &85.2 &\bf{85.3} &85.0 &84.8\\
				&Future Action Prediction &78.2 &78.3 &78.8 &78.7 &78.3
				&78.6 &79.1 &78.7 &78.8 &77.8\\
				\midrule
				\multirow{2}*{HDD} &Online Action Detection &\bf{45.1} &44.2 &43.8 &43.8 &42.9
				&42.0 &42.8 &\bf{45.1} &43.5 &43.2 \\
				&Future Action Prediction  &32.6 &32.6 &32.5 &32.0 &31.2
				&30.8 &30.9 &32.6 &32.2 &31.8 \\
				\bottomrule
		\end{tabular}}
		\caption{Performance evaluation at different temporal range size $K$ and action progression state space $P$ on the THUMOS-14~\cite{THUMOS_14}, TVSeries~\cite{tvseries} and HDD~\cite{HDD_dataset}.}
		\label{table:ablation_study_progression_num}
	\end{table*}
	
	\begin{figure*}[t]
		\centering
		\includegraphics[width=0.99\textwidth]{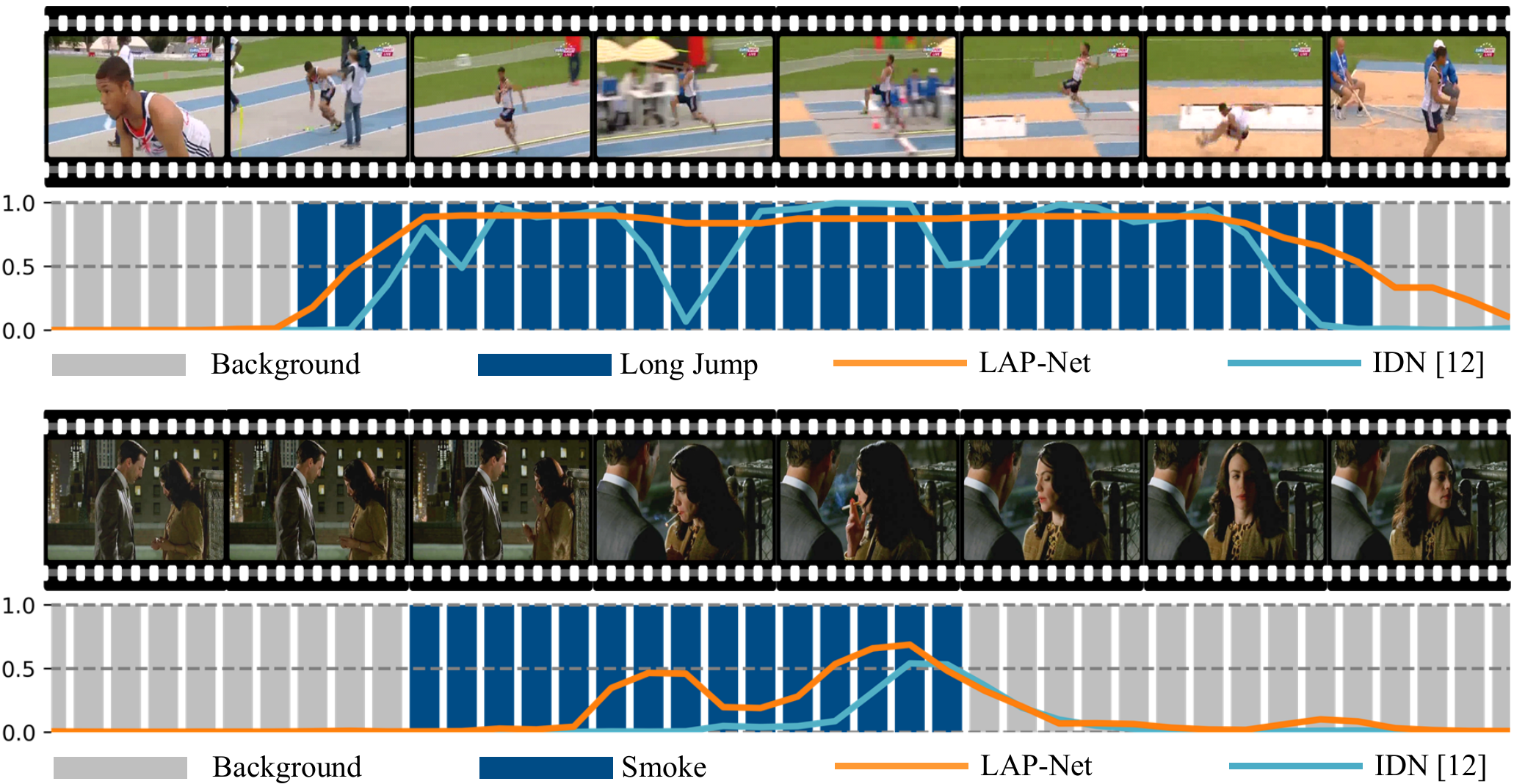}
		\caption{Qualitative result visualization of our LAP-Net and current state-of-the-arts IDN~\cite{IDU_2020} on THUMOS-14 and TVSeries Datasets. (Best viewd in color.)}
		\label{fig:qualitative_visualization}
	\end{figure*}
	
	\noindent \textbf{Effectiveness of Our Adaptive Features Sampling.}
	To evaluate the effectiveness of our adaptive supplementary features sampling strategy, we conduct an ablation study on THUMOS-14, TVSeries and HDD datasets. The online action detection results are reported in Table~\ref{table:ablation_study_RNN}. We denote our adaptive supplementary features sampling strategy as AFS. We can see that the introduction of our adaptive features sampling strategy can significantly boost the online action detection performance through the results.
	\\
	\\
	\noindent \textbf{Study on Temporal Range Size.} To explore the optimal temporal range size $K$ for obtaining supplementary representations, we make an extensive ablation study on THUMOS-14, TVSeries, and HDD datasets. We explored and evaluated the temporal range size $K \in \{3, 5, 7, 9, 11\}$. and reported the results in Table~\ref{table:ablation_study_progression_num}.
	This table also presents the ablation study results about the action progression stage, which we will study in the next subsection.
	During the experiments, we set the progression state space $P = 4$. 
	According to the results, we can see that the optimal temporal range size $K$ is different for distinct datasets.
	Specifically, for THUMOS-14 dataset the optimal $K$ is seven, for the TVSeries dataset is nine and for the HDD dataset is three. 
	We attribute these differences to the action tempo distinctions among these datasets. \emph{e.g.} the HDD dataset is continuously recorded under autonomous driving scenes, the behavior or action of cars usually goes very quickly. A relatively small temporal range size may capture more related information. In contrast, the THUMOS-14 and TVSeries are recorded under sports or daily life scenes, these human actions are relatively slow. A relatively wide temporal range can accumulate more semantic information.
	\\
	\\
	\noindent \textbf{Study on Action Progression State.}
	To explore the optimal action progression state space for online action detection, we also make an extensive study about the action progression space $P \in \{2, 3, 4, 5, 6\}$ on THUMOS-14, TVSeries and the HDD datasets. 
	The results are also reported in Table~\ref{table:ablation_study_progression_num}. We can see that a large progression stage space does not guarantee a better performance. This is possibly because it is not easy for the LAP-Net to distinguish the ambiguous boundaries between different action progression with increasing action progression space.
	
	\subsection{Qualitative Results}
	\par For qualitative results evaluation, we present the online action recognition results on the THUMOS-14, TVSeries in Fig.~\ref{fig:qualitative_visualization}. With the introduction of our adaptive supplementary features sampling strategy, our LAP-Net can achieve more stable and accurate recognitions compared with IDN~\cite{IDU_2020}.
	\section{Conclusion}
	\par In this paper, we have investigated how action progression will affect online action detection and propose a novel online action detection network, \emph{Learning Action Progression Network (LAP-Net)}. The LAP-Net integrates an adaptive feature sampling strategy to adaptively aggregate the optimal supplementary features conditioned on current action progression. Extensive experiments on three action detection datasets demonstrate our \emph{Learning Action Progression Network (LAP-Net)} superiority over current state-of-the-art methods. For the future work, we will extend our idea to other video understanding tasks.
	
	{\small
		\bibliographystyle{ieee_fullname}
		\bibliography{lapnet_refs}
	}
\end{document}